\documentclass[letterpaper, 10 pt, journal, twoside]{ieeetran}

\IEEEoverridecommandlockouts                              % This command is only needed if 
\title{
Memory-Maze: Scenario Driven Visual Language Navigation Benchmark for Guiding Blind People
}

\author{
Masaki Kuribayashi$^{*1,2}$, Kohei Uehara$^{*2}$, Allan Wang$^{2,3}$, Daisuke Sato$^{3}$,\\ Renato Alexandre Ribeiro$^{2}$, Simon Chu$^{3}$, and Shigeo Morishima$^{1}$

\thanks{
This work was supported by Waseda Research Institute for Science and Engineering and JSPS KAKENHI (23KJ2048 and 21H05054).} \thanks{$^{*}$Masaki Kuribayashi and Kohei Uehara contributed equally to this work.} \thanks{$^{1}$Masaki Kuribayashi and Shigeo Morishima are with Waseda University, Japan ({\tt\footnotesize rugbykuribayashi@toki.waseda.jp} and {\tt\footnotesize shigeo@waseda.jp}). }
\thanks{$^{2}$Masaki Kuribayashi, Kohei Uehara, Allan Wang, and Renato Alexandre Ribeiro are with Miraikan - The National Museum of Emerging Science and Innovation, Japan ({\tt\footnotesize\{ masaki.kuribayashi, kouhei.uehara, allan.wang, renato.ribeiro\}@jst.go.jp}).}
\thanks{$^{3}$Allan Wang, Daisuke Sato, and Simon Chu are with Carnegie Mellon University, United States ({\tt\footnotesize\{allanwan, daisuke, cchu2\}@andrew.cmu.edu}).} 
}

\usepackage{url}
\usepackage{graphicx}
\usepackage{multirow}
\usepackage{booktabs}
\usepackage{color}
\usepackage{amssymb}
\usepackage{newunicodechar}
\usepackage[utf8]{inputenc}
\usepackage[T1]{fontenc}    % use 8-bit T1 fonts
\usepackage{pifont}
\usepackage{newunicodechar}
\newunicodechar{✓}{{\ding{51}}}
\PassOptionsToPackage{hyphens}{url}\usepackage[hidelinks]{hyperref}

\usepackage{subcaption}
\usepackage{listings}
\usepackage{xcolor}
\begin{document}

\def\eg{{\it e.g.}}
\def\cf{{\it c.f.}}
\def\ie{{\it i.e.}}
\def\etal{{\it et al. }}
\def\etc{{\it etc}}

\newcommand{\red}{\textcolor[rgb]{0,0,0}}
\newcommand{\blue}{\textcolor[rgb]{0,0,0}}
\newcommand{\boldparagraph}[1]{\vspace{0.2cm}\noindent{\bf #1:}}
\newcommand{\revise}{\textcolor[rgb]{0.0,0.0,0.0}}

% \input{sections/response_letter}
% \clearpage
\maketitle

\begin{abstract}
Visual Language Navigation (VLN) powered robots have the potential to guide blind people by understanding route instructions provided by sighted passersby. This capability allows robots to operate in environments often unknown a prior. Existing VLN models are insufficient for the scenario of navigation guidance for blind people, as they need to understand routes described from human memory, which frequently contains stutters, errors, and omissions of details, as opposed to those obtained by thinking out loud, such as in the R2R dataset. However, existing benchmarks do not contain instructions obtained from human memory in natural environments. To this end, we present our benchmark, Memory-Maze, which simulates the scenario of seeking route instructions for guiding blind people. Our benchmark contains a maze-like structured virtual environment and novel route instruction data from human memory. Our analysis demonstrates that instruction data collected from memory was longer and contained more varied wording. We further demonstrate that addressing errors and ambiguities from memory-based instructions is challenging, by evaluating state-of-the-art models alongside our baseline model with modularized perception and controls.
\end{abstract}


\section{Introduction}
\IEEEPARstart{V}{isual} language navigation (VLN) is a task where an agent with visual access to the surroundings navigates under a human's instructions~\cite{anderson2018vision}.
Recently, navigation robots for blind people have been developed to help them gain independence~\cite{guerreiro2019cabot,liu2023dragon,kuribayashi2023pathfinder}, such as robots that allow users to choose destinations within prebuilt maps~\cite{guerreiro2019cabot,liu2023dragon}.
One scenario in which such robots would benefit from the VLN technology is where blind people request instructions to their destinations from sighted passersby in unfamiliar buildings~\cite{muller2022traveling}.
In this scenario, the VLN technology deployed on navigation robots may assist their blind users by understanding verbal instructions from the passersby and then autonomously guiding them to their destinations.
VLN technology could also allow robots to operate autonomously without relying on building infrastructure or prebuilt maps, which is crucial for allowing robots to assist blind people in navigating various new environments~\cite{kuribayashi2023pathfinder,kuribayashi2025wanderguide}.

\begin{figure}[t]
\centering
\includegraphics{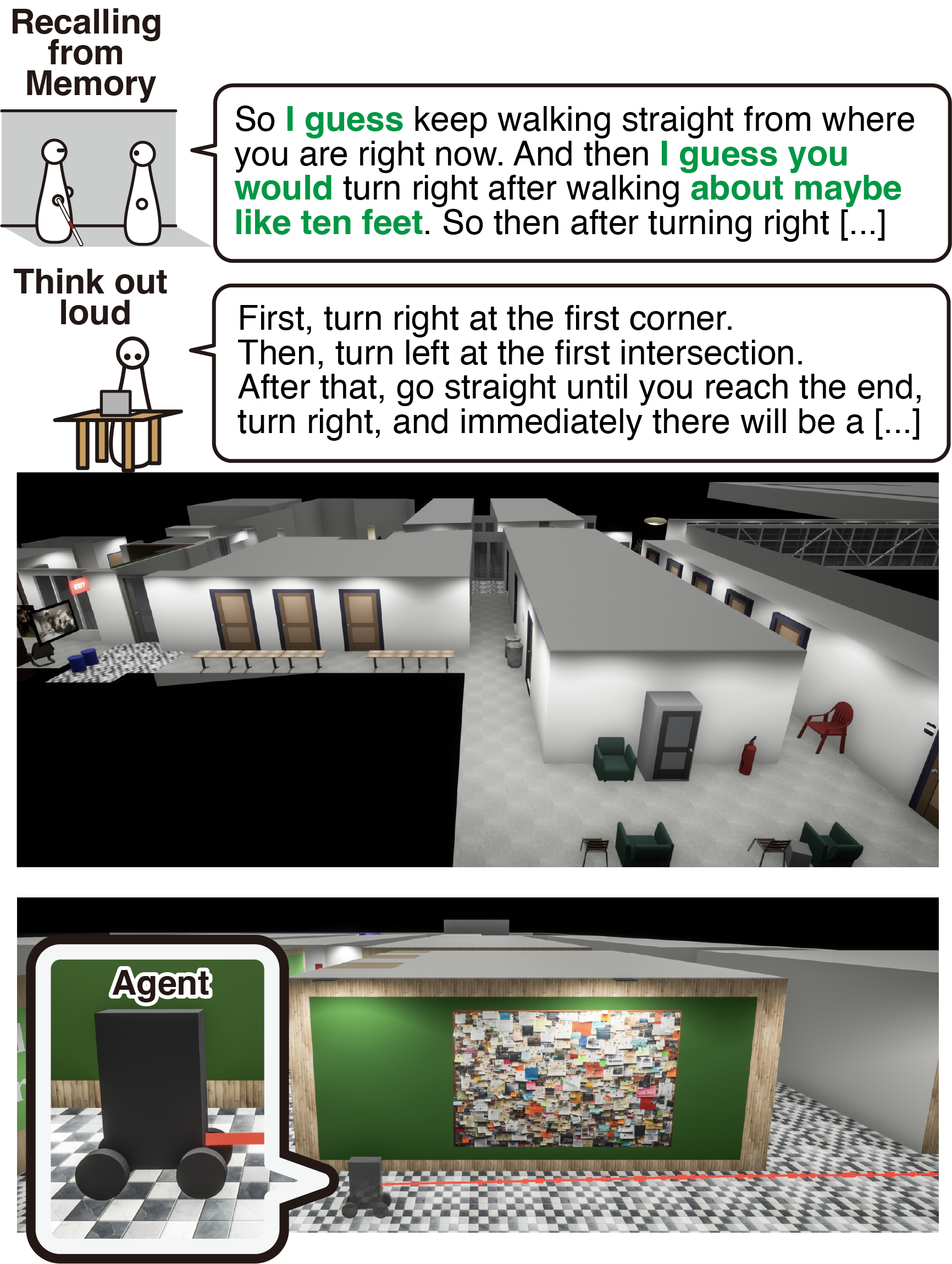}
\caption{\textbf{Memory-Maze Benchmark.} Top: the instructions obtained in the memory-based scenario contain unique phrases, highlighted in green, in contrast to those collected in traditional think-out-loud settings. Middle: Our benchmark environment based on the CARLA simulator~\cite{Dosovitskiy17}. Bottom: the VLN agent that navigates within the environment.}
\vspace{-2em} 
\label{fig:teaser}
\end{figure}

However, direct application of existing VLN models to the blind people navigation scenario is currently limited, as there is a need for a benchmark that reflects the blind users' demands realistically.
Many VLN tasks have been addressed in environments such as static houses~\cite{anderson2018vision} or roadways~\cite{chen2019touchdown}. 
Nonetheless, it is also most important for blind individuals to navigate large public spaces such as shopping malls or university hallways.
Compared to existing environments, these environments are characterized by physical turning points and intersections, resembling a maze.
Besides the environmental difference, in existing VLN literature, natural language instructions are provided by thinking out loud.
In other words, annotators visually navigate a virtual environment and type out instructions for constructing routes concurrently. 
In our scenario, sighted passersby must describe the route from their memory, which often contains errors such as inaccurate estimates of distances, hallucinations of landmark objects, and omissions of key turning points.
To the best of our knowledge, our benchmark is the first to address the scenario of a blind user seeking memory-based instructions from sighted passersby in maze-like public spaces.

We present \textit{Memory-Maze} (Fig.~\ref{fig:teaser}), a benchmark that reflects the blind user navigation scenario.
Memory-Maze contains virtual environments of real-world public spaces. 
It is based on CARLA~\cite{Dosovitskiy17}, which enables us to simulate various sensor data (\eg, LiDAR) from robots.
It also contains instructions data gathered from two studies from sighted individuals.
In the first study, instructions were gathered through online questionnaires by observing walk-through videos from a first-person perspective. 
This is similar to the annotation method used in existing research. 
In the second study, instructions were collected in-person by asking sighted passersby to describe the same routes from their memories. 
This reflects the novel scenarios envisioned in our benchmark.
We observed different characteristics among the two studies in terms of length, number of errors, variety, among others.

\looseness=-1 
\red{To analyze the difficulty of our benchmark, we developed a VLN baseline model better designed to navigate in large public spaces, by leveraging modular APIs to handle navigation control and perceptions.
Our model also fulfills two requirements for the practical deployment of VLN models for blind people: zero-shot transfer to unseen environments without navigation graphs and single inference.
Navigation robots need to be used in unseen environments for blind people, directly applying existing supervised models poses a challenge due to their limited performance in unseen settings~\cite{wu2023vision}.
Additionally, existing models perform repeated iterative inferences during navigation, resulting in frequent stops and prolonged navigation time. 
Leveraging large language models' (LLM) potential for zero-shot generalization in unseen environments, our single-inference LLM-powered model converts the instruction into Python code based on the defined robot control API (Sec.~\ref{sec:api}) for route navigation.
This code generation approach modularizes low-level commands such as path-planning for collision avoidance and intersection detection, and serves as a baseline that focuses more on the language interpretation and reasoning capabilities of VLN.
Through the study with our model and the current state-of-the-art methods~\cite{zhou2023navgpt,zhang2024navid}, we demonstrated the difficulty of our benchmark and a tendency that real-world memory-based instructions are more difficult for VLN models to handle.}

We summarize our contributions below.
\begin{enumerate}
\item We constructed Memory-Maze, a benchmark containing virtual environments of a \red{large public spaces}, and gathered two sets of instructions, one collected by thinking out loud and one obtained from human memory.
\item \red{Through an experiment with current state-of-the-art models and our baseline VLN model, we revealed the gap between the instructions collected based on memory and those collected by thinking out loud.}
\end{enumerate}

\revise{
Our benchmark and codes are available at \href{https://github.com/chestnutforestlabo/MemoryMaze}{https://github.com/chestnutforestlabo/MemoryMaze}
% \url{https://github.com/chestnutforestlabo/MemoryMaze}
}
\section{Related Work}

\begin{figure*}[t]
  \centering
  \begin{subfigure}[t]{0.33\linewidth}
    \centering
    \includegraphics[width=\linewidth]{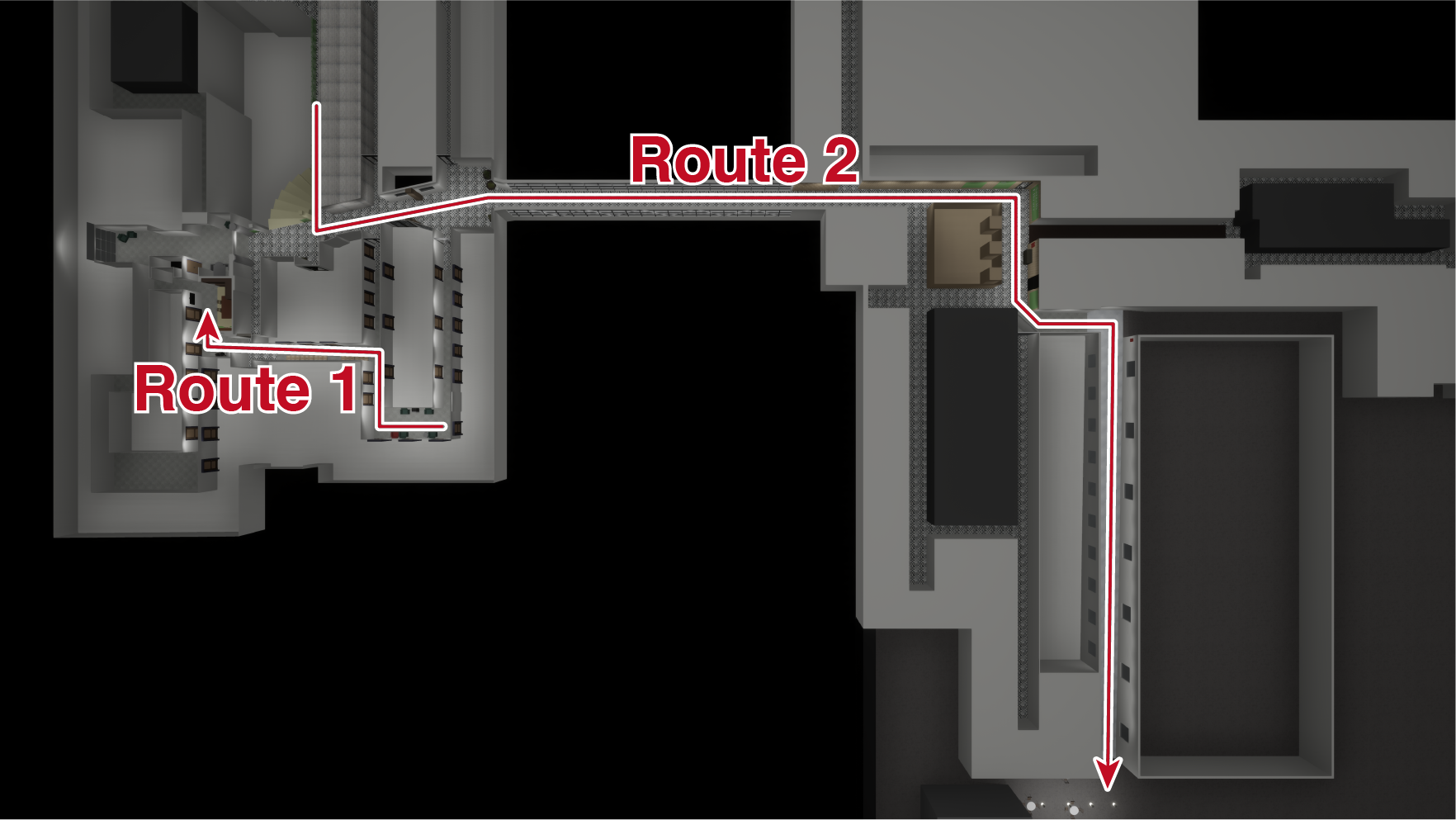}
    \caption{University}
  \end{subfigure}%
  % \hfill
  \begin{subfigure}[t]{0.33\linewidth}
    \centering
    \includegraphics[width=\linewidth]{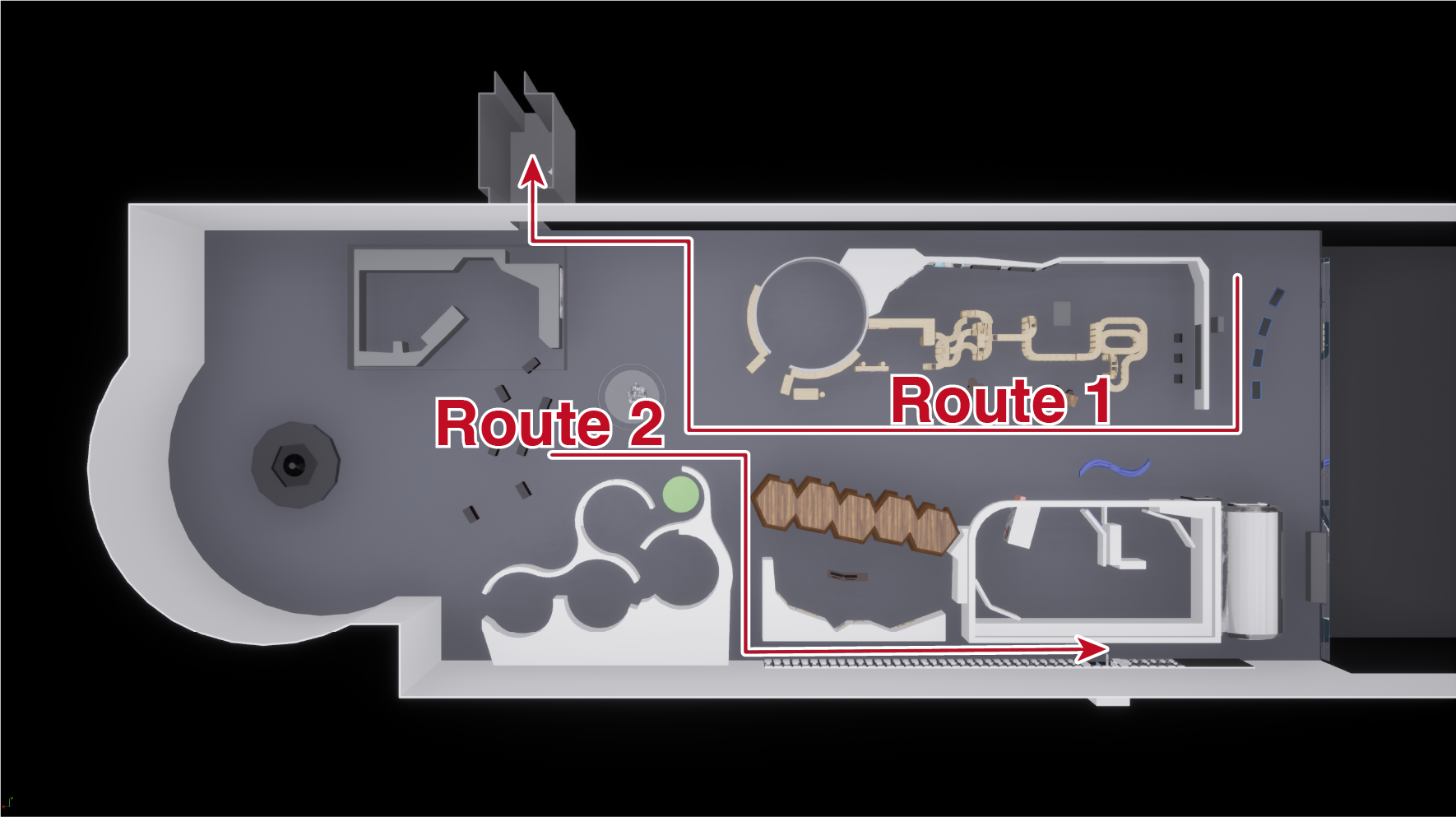}
    \caption{Museum 5F}
  \end{subfigure}%
  % \hfill
  \begin{subfigure}[t]{0.33\linewidth}
    \centering
    \includegraphics[width=\linewidth]{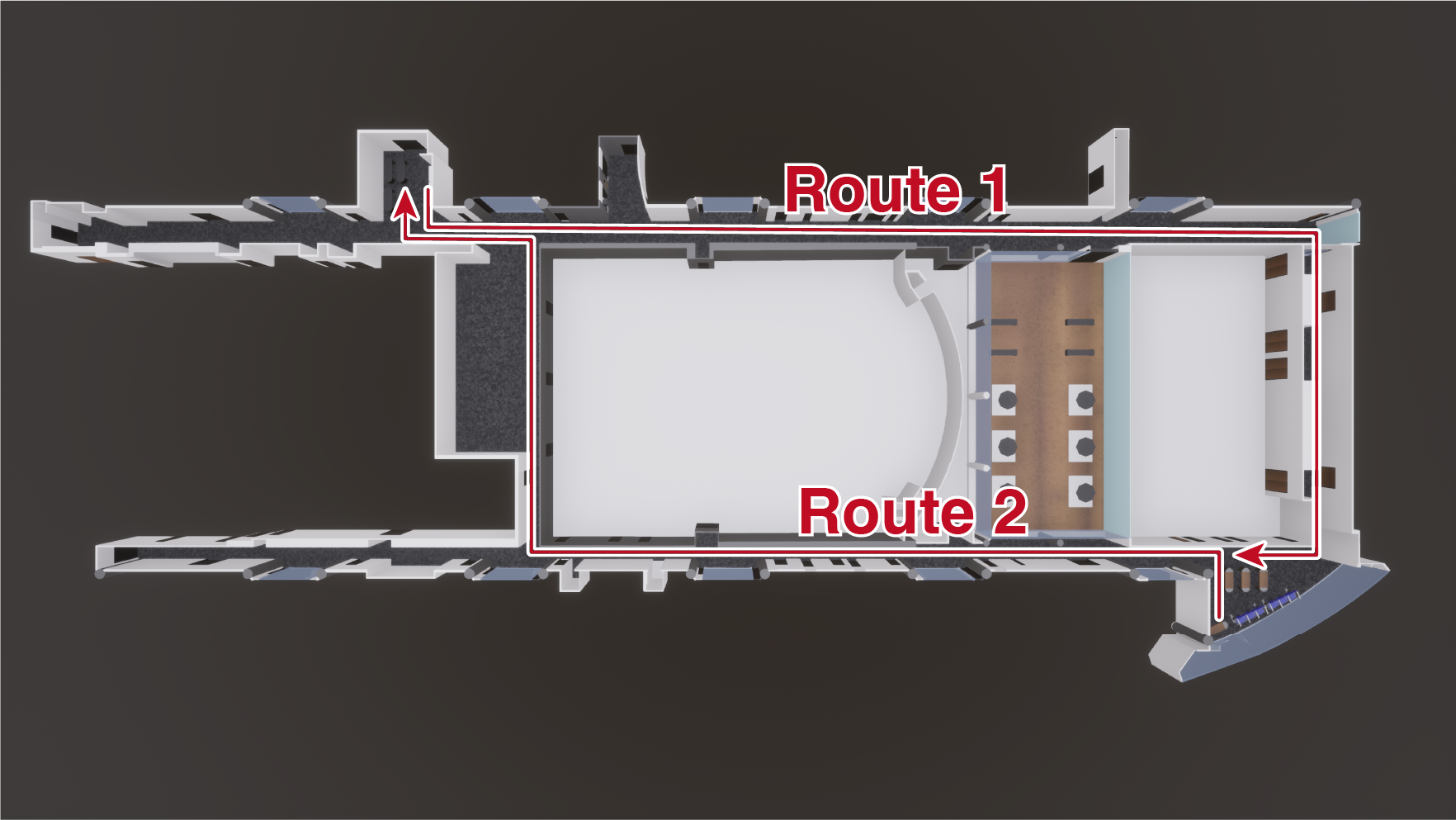}
    \caption{Museum 7F}
  \end{subfigure}
  \caption{\textbf{Bird's-Eye Views of Memory-Maze.} The benchmark contains three environments. The university includes features such as classrooms, offices, hallways, a kitchen, and a library. The 5th floor of the museum mainly contains exhibits. The 7th floor contains conference rooms, hallways, and a terrace area. Each environment includes two routes, totaling six routes. \red{In the on-site study, participants were asked to describe the route from the starting point to the end point, thus, their descriptions may vary from the visualized route.}}
  \label{fig:routeMap}
  \vspace{-1em}
\end{figure*}

\subsection{Assistive Navigation Systems for Blind People}
\label{sec:related1}
Recently, navigation robots have been explored to aid blind people in avoiding obstacles while navigating.
A common practice is to prepare prebuilt maps and infrastructure for localization and manual destination selection~\cite{guerreiro2019cabot,liu2023dragon}.
This practice poses a limitation for these systems, as prebuilt maps and infrastructure are costly to obtain and maintain.
Consequently, a map-less approach was also proposed~\cite{kuribayashi2023pathfinder,ranganeni2023exploring}.
\blue{For example, the PathFinder~\cite{kuribayashi2023pathfinder} system allows users to input navigation directions through the buttons on the robot’s handle based on instructions from sighted passersby~\cite{muller2022traveling}.}
However, because the system needs users to understand and memorize the instructions, high cognitive loads are placed on the users.
\red{To address this, we present a novel, practical benchmark for VLN models that aims to interpret memory-based instructions from sighted passersby and navigate users autonomously to their destinations.
}

\subsection{Benchmarks in VLN tasks}
\label{sec:relatedbenchmark}
The VLN task has been conducted in various benchmarks, ranging from indoor~\cite{anderson2018vision,qi2020reverie} to outdoor~\cite{chen2019touchdown} settings.
Most of the instruction annotations of these benchmarks were created by annotators who typed while concurrently observing a virtual environment or by researchers who constructed them manually.
This way of obtaining instructions is not suitable for our purpose, as it does not reflect the scenario of people describing routes from their memories.
\red{Researchers have also explored benchmarks with longer routes for long-horizon navigation tasks~\cite{song2025towards}.}
\red{Still, existing benchmarks} do not feature large public areas where blind people navigate, such as shopping malls or university hallways. 
These areas contain both static and dynamic obstacles and are characterized by the existence of turning points and intersections (Fig.~\ref{fig:routeMap}). 
\blue{A related benchmark is Touchdown~\cite{chen2019touchdown}, which also emphasizes navigation through intersections and dynamic environments.}
However, its map structure is represented by a navigation graph (\ie, an undirected graph that represents navigable points with nodes)\red{, whereas the \revise{Memory-Maze} assumes no prior knowledge such as navigation graphs.
}

\subsection{VLN Models}
Researchers have explored solutions for VLN tasks using supervised models~\cite{anderson2018vision}, which learn from a sequence of observations and actions to take.
These supervised models often do not transfer well in unseen environments~\cite{wu2023vision}.
With the recent advancements in LLM, researchers have also explored methods that do not require retraining~\cite{zhou2023navgpt,shah2023lm,huang2022language}.
One such approach was to use LLMs to extract landmarks from instructions and follow chronologically~\cite{shah2023lm}.
Another approach was to utilize LLM to flexibly determine actions at each step. 
NavGPT~\cite{zhou2023navgpt} is a model that uses LLM iteratively to select the node to navigate to within a navigation graph.
Additionally, researchers have explored approaches that utilize the code generation capability of LLM~\cite{huang2023visual,biggie2023tell}.
In the method proposed by Biggie~\etal~\cite{biggie2023tell}, given a prebuilt 3D map, images from their robot, and a Python API, the model generates codes that locate a target object~\cite{suris2023vipergpt}, maps the object's location on the 3D map, and navigates to the mapped location.
While these methods are effective when the given instructions include sufficient landmarks, instructions recalled from memory often contain insufficient landmarks, potentially leading to failure.
Furthermore, these methods are limited by the need for a navigation graph or 3D map, which is difficult to construct for every unseen environment.
To eliminate this requirement, models have been proposed to predict navigation graphs~\cite{krantz2021waypoint} or low-level actions~\cite{zhang2024navid} iteratively. 
However, the need for iterative inference prolongs inference time, which may affect navigation by not reacting to dynamic obstacles responsively.
\red{Moreover, iterative inference may be impractical in large public spaces, where the model could be required to perform over many inference steps, due to the need to process long time-horizon data.}
Our model utilizes LLM to produce navigation codes that follow a specified path in a single iteration, and allows flexible integration of low-level planning algorithms for obstacle avoidance.
This direct generation of navigation codes, coupled with existing low-level planning algorithms, allows operation without navigation graphs.
\section{Memory-Maze}

Here, we describe our benchmark's virtual environment and the robot simulation program.
To simulate our scenario, we selected a floor of a university building and two floors in a museum building (Fig.~\ref{fig:routeMap}), which is characterized by the existence of multiple turning points.

\subsection{Selecting and Building the Simulator}
To simulate a scenario where a robot guides a blind person, it is necessary to simulate high-fidelity egocentric visuals that are realistic enough to run an image recognition algorithm.
Thus, we built a novel virtual environment from scratch on top of the CARLA~\cite{Dosovitskiy17} simulator.
While primarily developed for autonomous driving simulations, CARLA's flexibility and compatibility with the Unreal Engine allowed us to create a detailed 3D model of the experimental site.
CARLA also offers the ability to configure the existence of static and dynamic obstacles and to simulate various sensors like RGB cameras, depth sensors, and LiDAR sensors.
We created a 3D model of the experimental site using Fusion 360 and imported it into CARLA.
This 3D model accurately reproduces the experimental site, both visually and in terms of floor layout.
It also includes major objects along the route (doors, chairs, a statue, \etc).

\looseness=-1 
\subsection{Implementation of the Control Program}
\label{sec:api}
Our next step was to develop a control program for the robot in the simulator \red{to be used by our baseline VLN model}.
Utilizing CARLA's Python API to control the navigation robot, we implemented various control functions.
We describe four major functions implemented.

We implemented functions for the agent to move forward (\texttt{move\_forward(distance)}), find a turning point (\texttt{detect\_turning\_point()}), and turn (\texttt{turn(direction)}) using CARLA's \texttt{vehicle.apply\_control} API.
When using the \texttt{move\_forward(distance)} function, to ensure the robot moves along the path without colliding with walls, we implemented a feature that makes the robot navigate as closely to the center of the corridor as possible.
We calculate the central path based on the coordinates of the four corners of the corridor in the 3D model.
The central path tracking is realized through PID control, which adjusts the robot's steering angles.
When the \texttt{detect\_turning\_point()} function is used, it determines if the robot is in the pre-annotated areas of turning points and returns navigable directions if the robot is in one of them.
Once the robot is at the turning point, it could change its direction using the \texttt{turn(direction)} function.
\red{Because component algorithm development of the control program was beyond the scope of this study,} coordinates of the corridor's corners and the turning point areas are acquired from the virtual environment, reducing errors from noise in perception or control, and focusing on executing instructions.
However, these can be obtained using prior methods~\cite{kuribayashi2023pathfinder}.

Additionally, we implemented an image recognition module \texttt{detect\_from\_RGB\_image(object)}, \red{which outputs bounding boxes of all detected objects,} to manage landmark-related instructions such as \textit{``turn after finding a chair.''} 
While most existing object detection models are designed to identify objects from predefined classes, they are not capable of detecting arbitrary objects.
Therefore, we used Grounding DINO~\cite{liu2023grounding}, an open-vocabulary object detection model.
Open-vocabulary object detection models output bounding boxes for any object by using the object's name as a query.
With the object detection model selected, we then used CARLA's robot ego-centric RGB sensors to capture images.
To address tasks requiring the robot to identify an object multiple times (\eg, \textit{ ``turn after passing four doors''}), we added tracking algorithms to avoid counting the same object in different frames as distinct entities.
We further assume that in instructions that require finding landmark objects, the objects are located in close vicinity.
For example, in the instruction \textit{``turn after finding [object],''} the camera may capture the object at a considerable distance, but such instructions typically mean the object is close to the robot. % although the camera could capture the object at a considerable distance, such instructions typically imply that \textit{``[object]''} is near the robot.
\revise{
 To achieve this, we used CARLA’s depth sensors to measure the distance to each detected object and filtered out those beyond four meters, ensuring that only nearby objects were considered. % Therefore, we used the depth sensors in CARLA to measure the distance to each object in the image, filtering out objects that are far away to ensure only those at close range are detected.
% Distance threshold is set to be four meters.
}

\begin{table}[t]
\centering
\caption{\textbf{Data Analysis.} The table presents the route length (RL), mean, median, and standard deviation (SD) of word counts in the collected instructions, and their failure rates (FR). 
For the onsite instructions, we also report the alternative rate (AR), the rate of describing alternative routes.
}
\label{tab:analysis}
\resizebox{\columnwidth}{!}{%
\begin{tabular}{clccccccc}
\toprule
\multicolumn{1}{l}{}          & \multicolumn{1}{l}{Route} & RL & Iteration  & Mean  & Median & SD & FR & AR\\
\midrule
\multirow{12}{*}{\begin{tabular}{c}
\revise{Online} \\
\revise{Think-Out-Loud} \\
\revise{Instruction}
\end{tabular}}   & \multirow{2}{*}{University R1} & \multirow{2}{*}{40.27m} & 1          & 51.8  & 47.0   & 17.8  & 0.0\%  & -            \\
                                 &                                &                        & 2          & 69.8  & 64.0   & 19.9  & 0.0\%   &-           \\
                                \cmidrule(lr){2-9} 
                                & \multirow{2}{*}{University R2} & \multirow{2}{*}{156.68m}  & 1          & 81.3  & 81.0   & 24.9  & 9.09\% &-               \\
                                &                                &                          & 2          & 98.3  & 99.0   & 31.2  & 3.03\%  &-             \\
                                \cmidrule(lr){2-9} 
                                & \multirow{2}{*}{Museum 5F R1} & \multirow{2}{*}{71.18m} &1          & 81.4 & 78.0 & 36.3 & 17.39\% &-\\ %TO UPDATE
                                &                               &                        &2          & 88.9 & 90.0 & 32.3 & 17.39\% &-\\ %TO UPDATE
                                \cmidrule(lr){2-9} 
                                & \multirow{2}{*}{Museum 5F R2} & \multirow{2}{*}{44.05m}  & 1          & 60.1 & 53.5 & 21.5 & 9.09\% &-\\
                                &                               &                         & 2          & 71.0 & 61.0 & 33.9 & 4.55\% &-\\
                                \cmidrule(lr){2-9} 
                                & \multirow{2}{*}{Museum 7F R1} & \multirow{2}{*}{86.10m} &1          & 98.2 & 91.5 & 42.5 & 13.64\% &-\\
                                &                               &                        &2          & 96.7 & 90.0 & 42.2 & 18.18\% &-\\
                                \cmidrule(lr){2-9} 
                                & \multirow{2}{*}{Museum 7F R2} & \multirow{2}{*}{79.40m} & 1          & 71.3 & 68.0 & 25.8 & 4.35\% &-\\ %TO UPDATE
                                &                               &                        &2          & 95.0 & 85.0 & 47.4 & 0.00\% &-\\ %TO UPDATE
                                \midrule
\multirow{12}{*}{\begin{tabular}{c}
\revise{Onsite} \\
\revise{Memory-Based} \\
\revise{Instructions}
\end{tabular}}   & \multirow{2}{*}{University R1} & \multirow{2}{*}{40.27m} & 1          & 73.9  & 74.5   & 36.6  & 25.0\% & 10.0\%             \\
                                 &                                &                        &2          & 102.9 & 94.5   & 51.1  & 25.0\%   & 10.0\%            \\
                                \cmidrule(lr){2-9} 
                                & \multirow{2}{*}{University R2} & \multirow{2}{*}{156.68m}  & 1          & 131.0 & 115.5  & 73.2  & 40.0\%   & 15.0\%           \\
                                &                                &                        &2          & 147.3 & 143.0  & 65.0  & 35.0\%     & 15.0\%         \\
                                \cmidrule(lr){2-9} 
                                & \multirow{2}{*}{Museum 5F R1} & \multirow{2}{*}{71.18m} & 1          & 68.2 & 64.0 & 27.4 & 76.19\% & 0.0\%\\
                                &                               &                        &2          & 97.1 & 92.0 & 27.0 & 9.52\% & 0.0\% \\
                                \cmidrule(lr){2-9} 
                                & \multirow{2}{*}{Museum 5F R2} & \multirow{2}{*}{44.05m}  & 1          & 65.5 & 51.0 & 42.7 & 45.45\% & 59.1\%\\
                                &                               &                        &2          & 83.4 & 68.5 & 39.7 & 4.55\% & 63.6\%\\
                                \cmidrule(lr){2-9} 
                                & \multirow{2}{*}{Museum 7F R1} & \multirow{2}{*}{86.10m} &1          & 68.7 & 69.5 & 27.5 & 54.54\% & 4.5\%\\
                                &                               &                        &2          & 89.0 & 84.0 & 24.0 & 13.64\% & 9.0\%\\
                                \cmidrule(lr){2-9} 
                                & \multirow{2}{*}{Museum 7F R2} & \multirow{2}{*}{79.40m} &1          & 79.5 & 69.0 & 40.0 & 52.38\% & 85.7\%\\
                                &                               &                        &2          & 99.0 & 96.0 & 37.5 & 23.81\% & 90.1\%\\
\bottomrule
\end{tabular}%
}
\vspace{-1em}
\end{table}
\vspace{-1mm}
\section{Instruction Data Collection}

\subsection{Procedure}
We conducted two studies, one online and one onsite, to collect natural language instruction data for routes at three locations: a floor across three buildings in a university and two floors in a museum.
We designed the route as shown in Fig.~\ref{fig:routeMap}.
The studies were approved by our institutional review board (IRB), and informed consent was obtained from all participants.
For each route, we obtained two rounds of instructions: one asking participants to describe the route to a blind person with a navigation robot naturally (first iteration) and another asking participants to describe the route after providing them with a brief description of the capability of the navigation robot (second iteration). 
The second instruction was collected to obtain more accurate memory-based instructions given by passersby.
This was achieved by explaining the robot's capability (\eg, being able to detect objects) to the participants.
It simulates a scenario where a blind user may provide sighted passersby with robot information to obtain refined instructions.
We expect that telling them about robots' capabilities would enable VLN models to achieve better performance.

In the first study, participants completed an online questionnaire designed to gather instructions that were similar to those in prior works.
They were first presented with a scenario in which they communicated with a blind person accompanied by a navigation robot capable of following natural language instructions and 360$^{\circ}$ video walkthroughs of two routes.
They were then asked to type instructions to the destination. 
They were allowed to re-watch the walkthrough videos at any time.
We collected four instructions per participant.
In total, 78 participants participated in the study, resulting in 312 instructions.
% 33 CMU, 45 Miriakan
The participants were gathered through university recruitment or through an online survey platform, and all were unfamiliar with the shown routes.
The study was conducted in Japan, and the instructions were translated into English using GPT-4.

In the second study, we conducted an onsite in-person study.
The aim of this study was to gather data that reflects the realistic scenario of sighted people describing the route from their memory. 
Thus, they did not watch the walkthrough video or experience the route during the study.
The experimenter roleplayed as blind individuals, asked them for directions to the route destinations, and instructed them to describe the route verbally in two rounds.
For the first iteration, we asked participants to describe the routes as naturally as possible.
For the second iteration, to obtain more accurate instructions for the benchmark, in addition to explaining the robot's capabilities, the experimenter pointed out errors in the participants’ given instructions, such as a missing turn, and asked them to explain the route again.
For the university routes, we recruited sighted passersby and ensured that all participants were familiar with the route by using a pre-study check survey.
In this study, each participant described a single route, resulting in two instructions per participant.
In total, 40 participants participated in the study at the university, contributing 80 instructions.
For the museum routes, we recruited staff or recent visitors who were familiar with the museum layouts.
In this study, each participant described two routes, resulting in four instructions per participant.
In total, 43 participants participated in the study at the museum, contributing 172 instructions.

\begin{figure}[t]
  \centering
  % First row
  \begin{subfigure}{0.49\columnwidth}
    \includegraphics[width=\linewidth]{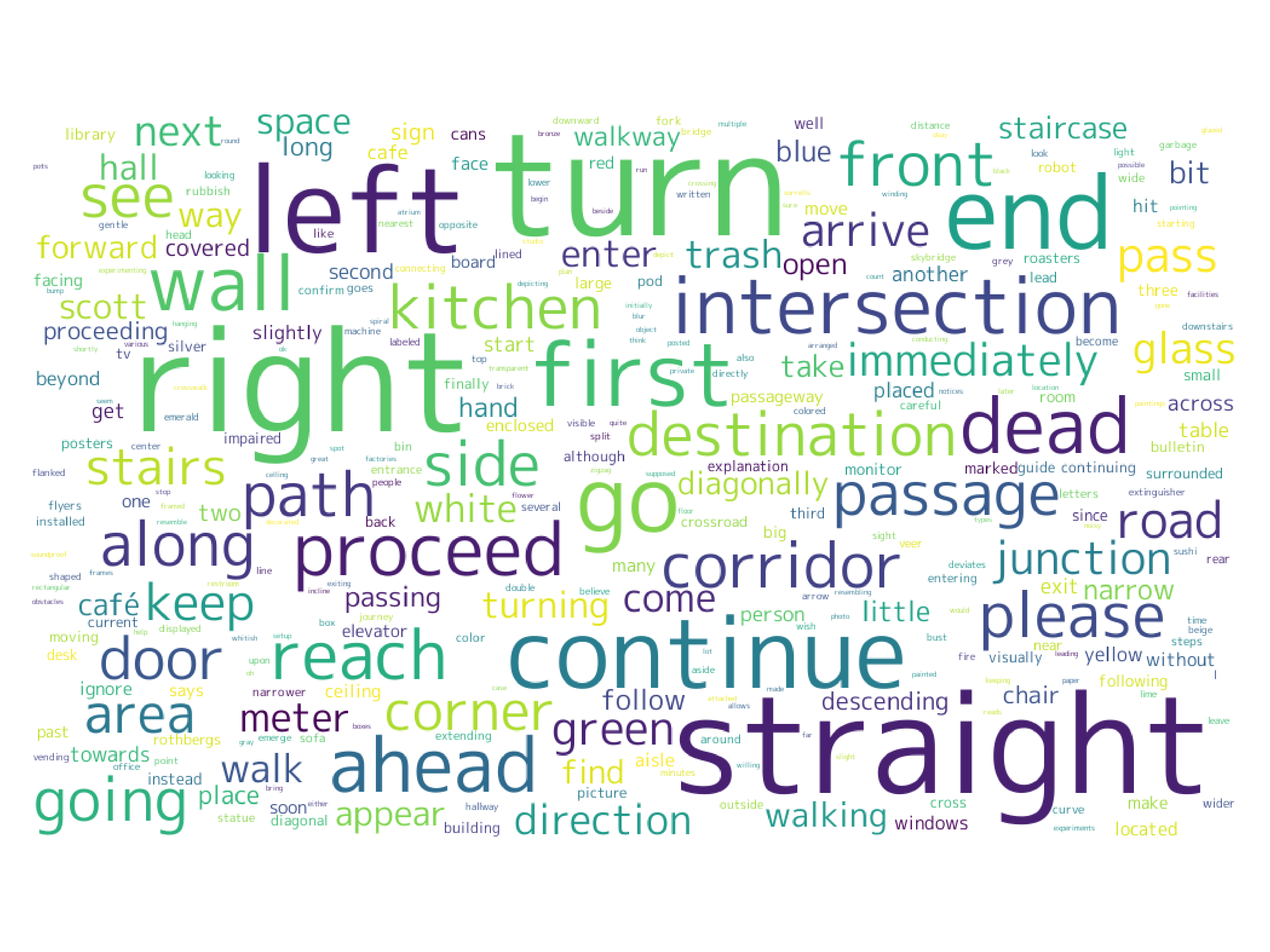}
    \vspace{-1cm}
    \caption{University Online Study}
  \end{subfigure}
  \hfill
  \begin{subfigure}{0.49\columnwidth}
    \includegraphics[width=\linewidth]{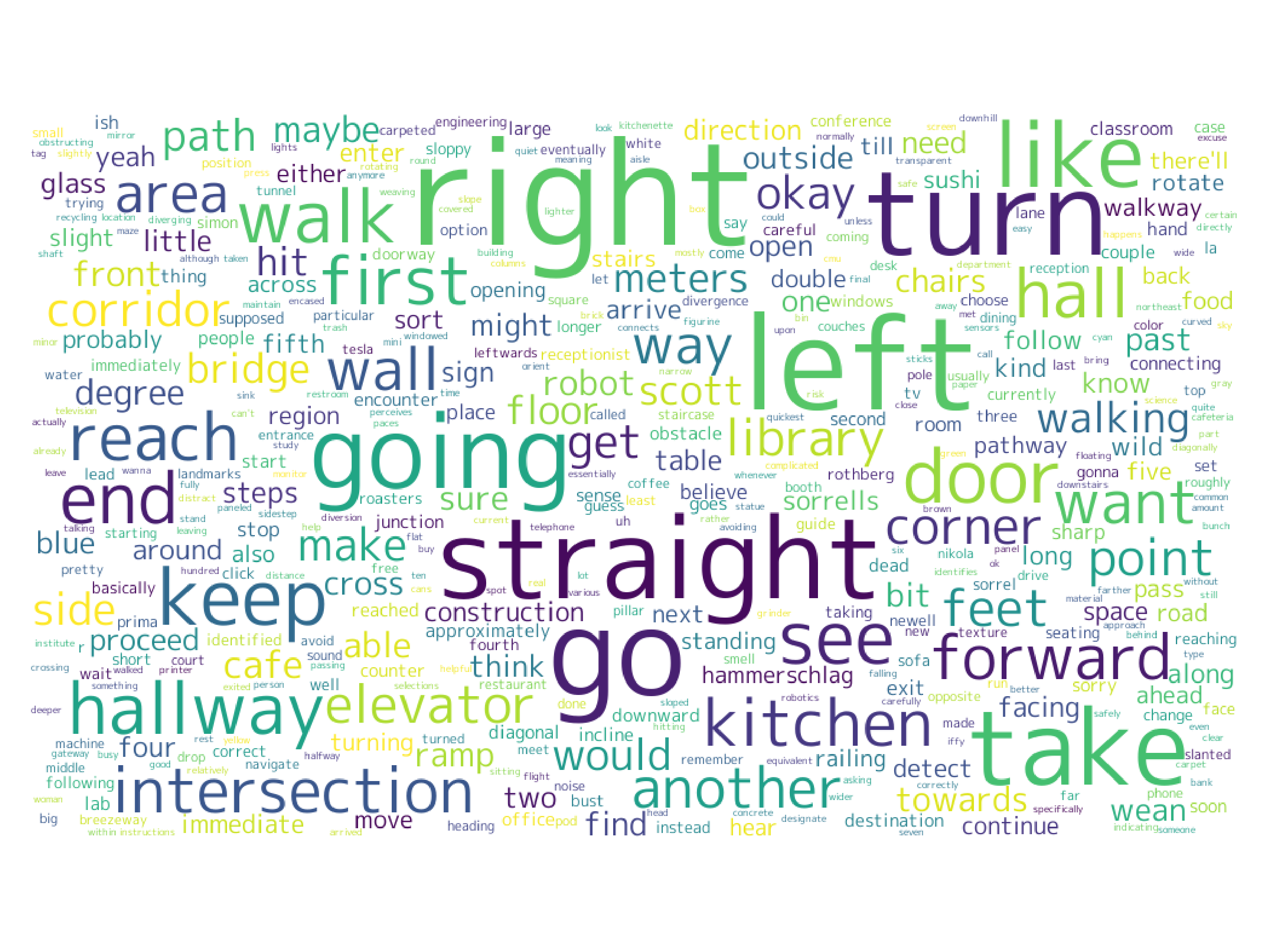}
    \vspace{-1cm}
    \caption{University Onsite Study}
  \end{subfigure}
  % Second row
  \begin{subfigure}{0.49\columnwidth}
    \includegraphics[width=\linewidth]{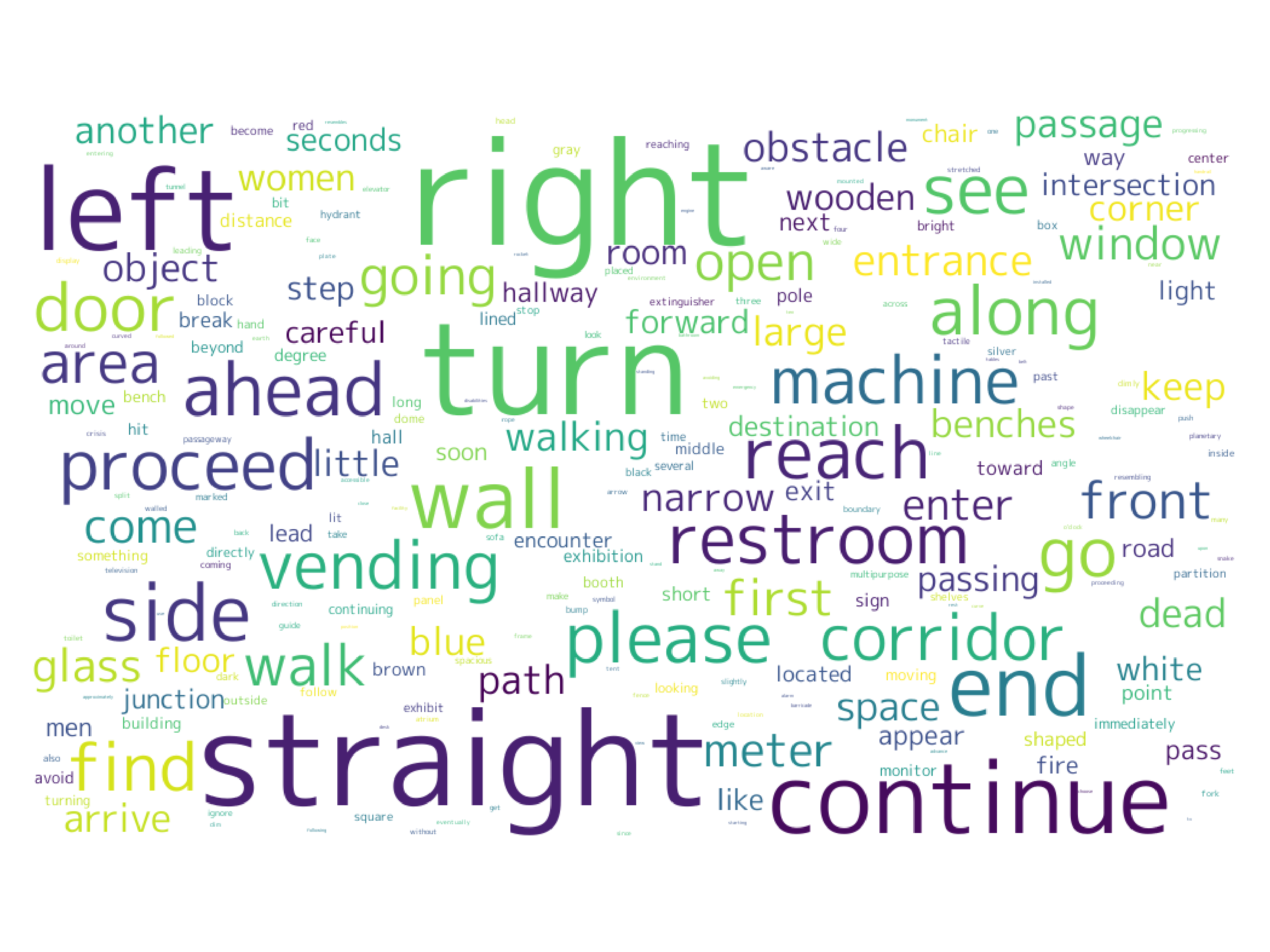}
    \vspace{-1cm}
    \caption{Museum Online Study}
  \end{subfigure}
  \hfill
  \begin{subfigure}{0.49\columnwidth}
    \includegraphics[width=\linewidth]{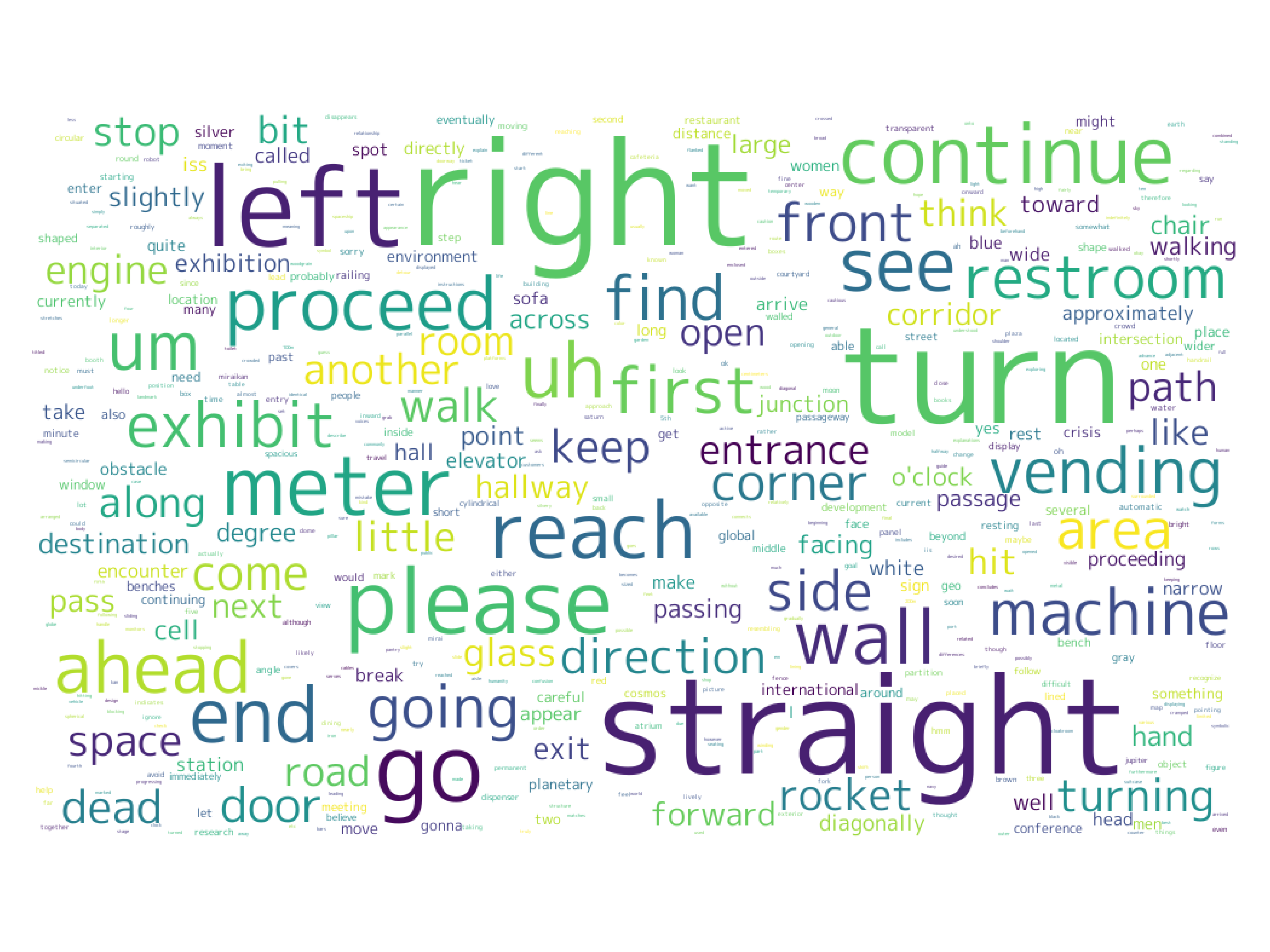}
    \vspace{-1cm}
    \caption{Museum Onsite Study}
  \end{subfigure}
  \caption{\textbf{Word Clouds.} The onsite instruction data contains unique phrases that come from talking while recalling from the memory, such as \textit{``uh,''} \textit{``maybe,''} and \textit{``okay.''}}
  \label{fig:cloud}
  \vspace{-1.5em}
\end{figure}

\subsection{Benchmark Analysis and Statistics}
\label{sec:analysis}
The mean, median, and standard deviation (SD) for the length of collected instructions are reported in Table~\ref{tab:analysis}.
First, we observed that the mean length and SD are longer for the second iteration in most cases, as participants tended to add more information on the second iteration.
Also, we observed a tendency for instructions collected onsite to have higher lengths and more length variation.
This is because, in the online study, participants described relevant and mostly accurate information about landmarks and turning points, while in the onsite study, many participants tried to be descriptive, relying on their memory, such as adding audio, olfactory cues, and conversational phrases such as \textit{``I'm not 100\% sure about this, but I think...''}.

The average instruction length and route distance in our benchmark are greater than those in previous datasets. 
For example, the R2R dataset includes instructions averaging approximately 30 words and route distances of about 10 meters~\cite{anderson2018vision}, and the RxR dataset features instructions averaging 78 words and route distances of 14.9 meters~\cite{ku2020room}. 

The word clouds of the collected instructions are shown in Fig.~\ref{fig:cloud}.
For the university environment, although the samples collected in the onsite study are fewer, they include 521 different words compared to the 381 words found in the samples from the online study. 
The same trend was noted in the museum environment, with 611 different words found in the onsite study and 586 words in the online study. 
This shows the greater diversity in the instructions' wording when described from memory.
Although the instructions from the online study were translated using LLM, we believe that these results hold in the instructions' original language.

In Table~\ref{tab:analysis}, we also manually analyzed each instruction to determine if it contained significant errors, \ie, the number of failures in describing the route correctly.
One author first conducted an initial failure review, after which multiple authors engaged in a discussion to reach a consensus on all samples.
\blue{\revise{Online think-out-loud instructions were} classified as failures for reasons such as turning in the wrong direction; instructing a turn at the incorrect turning point; and suggesting unnecessary extra turns.}
\revise{For onsite memory-based instructions,} the reasons for the failures were containing extra turns, directing to an incorrect direction, leading to a wrong destination, lacking essential turn information, turning to incorrect directions at a turn, and containing inaccurate environmental details. 

Interestingly, while examining the instructions, we realized that in the real world, humans may be able to correct errors in the instructions. For example, according to some passersby, the robot should go through a corridor between the hexagon exhibitions and the rectangular exhibition immediately to the right for museum 5F R2. However, there is actually no corridor between these two exhibitions. But it is possible to imagine where the nonexistent corridor might lead and try to find a detour. Being able to recognize errors in memory-based instructions is vital for aiding blind people to follow instructions provided by passersby. 
% This is a unique aspect of our benchmark.

As shown in Table~\ref{tab:analysis}, during the onsite study, we observed that participants sometimes described alternative routes compared to those we had initially anticipated and illustrated in Fig.~\ref{fig:routeMap}, when they described the routes from their memory.
Surprisingly, some participants described a different route in the second iteration compared to their initial description.
This highlights the potential of humans to describe alternative routes in real-world scenarios and the need for VLN models to perform equally well in these alternative routes, thereby underscoring the naturalness of our benchmark.

\red{
In summary, our benchmark captures real-world complexity beyond mere geometric turns.
We found that even routes with $90^\circ$ turns are accompanied by richly varied natural language instructions. 
For example, $\sim$25\% of participants described the two consecutive turns near the goal at Museum 5F R2 as \textit{``proceed in the front right direction''}.
In another example, one instruction described the consecutive turns in University R2 with only semantic cues such as
\textit{``[...], I believe and then you walk along the hallway connecting DEF hall and the GHI hall and then you'll arrive at your destination.''}
Memory-Maze provides a unique scenario-based evaluative environment by incorporating variation in intersection counts, route lengths, and landmark complexity. 
}

\begin{figure*}[t]
\centering
\includegraphics[width=\textwidth]{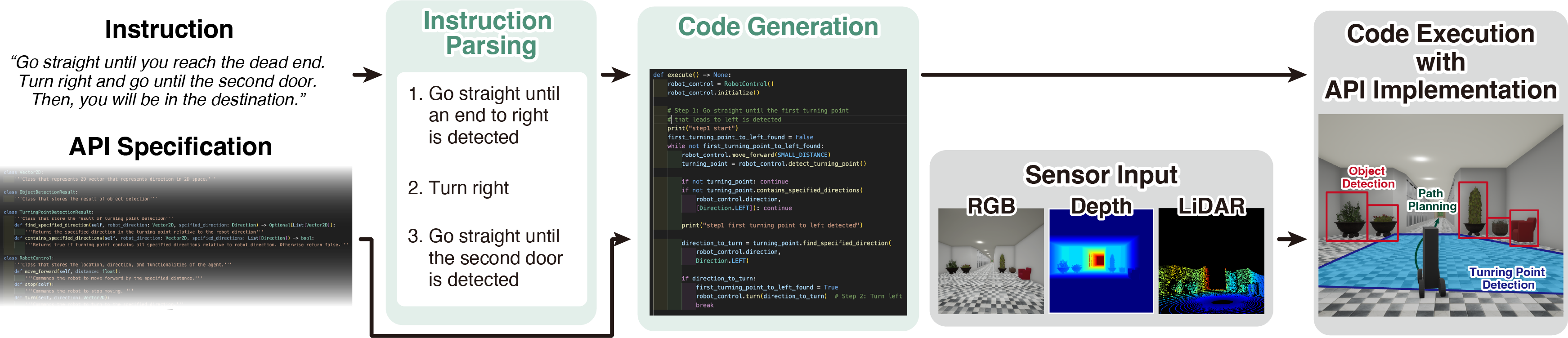}
\caption{\textbf{Method Overview.} Given a set of instructions from a sighted passerby, the LLM first parses it into an itemized format.
Then, combined with the API specification, the LLM generates Python code directly to control the robot, which runs in the virtual environment using the simulated sensor inputs.
}
\label{fig:implementation_overview}
\vspace{-1.5em}
\end{figure*}
\section{\red{Baseline} VLN Model Implementation}

\red{Our baseline VLN model uses the control API specification from our benchmark so that we may focus more on its language interpretation and reasoning capabilities}. 
% To satisfy the requirements of our scenario, it contains two characteristics. 
First, we utilized LLM's capability to generalize to various tasks and comprehend complex natural language instructions, so that no additional training is required when deployed in a new environment.
Second, our method requires only a single inference iteration to generate low-level navigation code for robot control, in contrast to existing models that \red{perform multiple} inferences during navigation, which may prolong navigation time. 
It also eliminates the need for a navigation graph by generating codes that directly interface with low-level navigation modules.
The generation of navigation code potentially leads to the flexibility of integrating existing, well-established methods into various modules, such as for obstacle avoidance~\cite{guerreiro2019cabot} or turning point detection~\cite{kuribayashi2023pathfinder,kuribayashi2025wanderguide}.

We define the following as inputs to the agent: \textit{natural language instruction}, the \textit{sensor input} which includes the details obtained from sensors, \textit{API specification} which consists of the commands and their explanations in Python that the agent can use as described in Sec.~\ref{sec:api}, \textit{API implementation} which is the actual implementation of the API specification, and the \textit{initial orientation} of the robot.
We assume the initial orientation is predetermined, as the blind user can adjust it in place.
We used the GPT-4 \red{(gpt-4-1106-preview)} model for the LLM.
For the initial setup of the prompt, instructions from five participants in the online study were used as references to construct the prompt for the proposed systems.
Fig.~\ref{fig:implementation_overview} \red{shows the implementation overview}.

\subsection{Parsing Instruction}
\label{sec:parsing}
The system first parses a natural language instruction to step-by-step instructions using LLM.
This was done to organize our benchmark's diverse natural language instruction and make it more interpretable before generating navigation codes.
To achieve this, we prompt LLMs with a set of rules they should follow, such as the requirement to describe when and which turning point to turn, and which object the robot should detect, along with examples of possible input and expected output.
After parsing, each navigation step is returned as a brief sentence.
We employ a two-stage prompting \red{method to guide LLM for more} accurate outputs. 
We prompt LLM to provide a thought to guide the generation of the first output, then refine the output by incorporating a second thought, leading to the finalized output.

\subsection{Navigation Code Generation}
\label{sec:navcodegen}
\red{To generate the navigation code,} we prompt LLM with an API specification that includes a range of commands for robot operations (\eg, \texttt{move\_forward(distance)} function). 
These commands are complete with docstrings of their usage explanation~\cite{biggie2023tell,suris2023vipergpt} and instructions to generate Python codes that follow the provided specification. 
For \texttt{detect\_from\_RGB\_image(object)}, our model uses an open vocabulary object detector internally (Sec.~\ref{sec:api}) and flexibly determines which object to detect by generating an object argument.
\red{For example, for an input requesting the location of a red chair, the function would be invoked as: \texttt{detect\_from\_RGB\_image("red chair")} by an LLM while generating a code.}
The API specification was formatted to the similar format of the previous work~\cite{biggie2023tell,suris2023vipergpt}, but with additional notes, such as how each function should be and not be used.
We \red{again} employ the same two-stage \red{prompting} method.
Finally, we execute the code using the API implementation.

\section{Experiment}
\label{sec:experiment}
\red{Our benchmark simulates a scenario in which blind people ask sighted passersby to provide route guidance~\cite{kuribayashi2023pathfinder} from their memories. 
To evaluate how current models perform under this setting, we conducted an experiment.}

\subsection{\red{State-of-the-Art Models}}
\looseness=-1
In our scenario, VLN agents are expected to demonstrate strong transferability, as blind users may navigate across diverse unseen locations by asking sighted passersby for directions.
To evaluate this capability, we compare our model with two prior state-of-the-art methods that leverage foundation models and exhibit strong zero-shot performance: NavGPT~\cite{zhou2023navgpt} and NaVid~\cite{zhang2024navid}. 

\textbf{NavGPT}~\cite{zhou2023navgpt} demonstrates strong zero-shot transfer capability by leveraging an LLM, visual foundation model, and an object detector to iteratively select destinations within a navigation graph until the agent determines it has reached the goal.
We used GPT-4o-mini for the LLM and for the visual foundation model, and the same Grounding DINO~\cite{liu2023grounding} for the object detector.
As NavGPT requires a navigation graph to operate, we constructed navigation graphs over the environments following the R2R dataset~\cite{anderson2018vision}. 

\textbf{NaVid}~\cite{zhang2024navid}, a state-of-the-art VLN model that demonstrates strong generalization to unseen environments, employs a visual foundation model and operates without a navigation graph, relying solely on camera input, similar to our model.
We strictly controlled the agent by following NaVid's established pipeline.
We initialized its weights \red{of LLM (Vicuna-7B)} using their open-sourced checkpoint~\cite{zhang2024navid}.

\vspace{-1mm}
\subsection{Metrics}
\revise{
For metrics}, we employ success rate (SR), oracle success rate (OSR), and shortest path distance (SPD)~\cite{anderson2018vision,gu2022vision}, and coverage weighted by length score (CLS)~\cite{jain2019stay,ilharco2019general}.
As CLS computes the similarity of the path on the graph, it requires a dense navigation graph to map the navigated trajectory onto.
Thus, we divided passable corridors into 50 cm square grids to serve as nodes on a graph and mapped predicted and ground truth paths onto it to calculate this metric~\cite{ilharco2019general}.
For routes where participants described an alternative path, we used the described route as the ground truth.

\section{Results and Discussion}
\label{sec:results}
\begin{table}[]
\centering
\caption{\textbf{Performance of VLN Models.} We compare our method with state-of-the-art VLN models that fulfill the requirements relevant to our scenario.}
\label{tab:result}
\resizebox{\columnwidth}{!}{%
\begin{tabular}{ccccccccccc}
\toprule
\multicolumn{3}{c}{Condition} & \multicolumn{4}{c}{\revise{Online Think-Out-Loud Instructions}} & \multicolumn{4}{c}{\revise{Onsite Memory-Based Instructions}} \\
Method  & Parser    & Route & SR↑   & OSR↑  & SPD↓    & CLS↑  & SR↑   & OSR↑  & SPD↓    & CLS↑  \\
  \cmidrule(lr){1-2} \cmidrule(lr){3-3} \cmidrule(lr){4-7} \cmidrule(lr){8-11} 
NavGPT                  &   & University R1     & 0.04 & 0.09 & 37.54 & 0.05 & 0.02 & 0.04 & 40.58 & 0.05 \\
NaVid                   &   & University R1     & 0.00 & 0.00 & 35.73 & 0.03 & 0.00 & 0.00 & 36.67 & 0.02 \\
Proposed                &   & University R1     & 0.20 & 0.24 & \textbf{17.01}  & \textbf{0.56} & 0.23 & 0.33 & 21.04  & 0.46 \\
Proposed                & ✓ & University R1     & \textbf{0.30 }& \textbf{0.35} & 19.66  & 0.49 & \textbf{0.30} & \textbf{0.38} & \textbf{18.32  }& \textbf{0.54} \\ 
\midrule 
NavGPT                  &   & University R2     & 0.00 & 0.00 & 162.10 & 0.01 & 0.00 & 0.00 & 161.13 & 0.01 \\
NaVid                   &   & University R2     & 0.00 & 0.00 & 149.79 & 0.00 & 0.00 & 0.00 & 151.47 & 0.00 \\
Proposed                &   & University R2     & 0.00 & 0.00 & 93.66  & 0.32 & \textbf{0.03} & \textbf{0.03} & 117.52 & 0.20 \\
Proposed                & ✓ & University R2     & \textbf{0.04} & \textbf{0.04 }& \textbf{81.59}  & \textbf{0.38} & \textbf{0.03} & \textbf{0.03} & \textbf{98.13}  & \textbf{0.29} \\
\midrule 
NavGPT                  &   & Museum 5F R1 & 0.00 & 0.00 & 50.76 & 0.00 &0.00 & 0.00 & 51.30 & 0.00 \\ %%TO BE UPDATED
NaVid                   &   & Museum 5F R1 & 0.00 & 0.00 & 54.59 & 0.01 & 0.00 & 0.00 & 55.50 & 0.01 \\ %TO BE UPDATED
Proposed                &   & Museum 5F R1 & 0.11 & 0.20 & 35.46 & 0.44 & 0.02 & 0.02 & 43.35 & 0.32 \\ %TO BE UPDATED
Proposed                & ✓ & Museum 5F R1 & \textbf{0.20} & \textbf{0.26} & \textbf{26.71} & \textbf{0.60} & \textbf{0.05} & \textbf{0.07} & \textbf{29.14} & \textbf{0.54} \\ %TO BE UPDATED
\midrule 
NavGPT                  &   & Museum 5F R2 & 0.00 & 0.07 & 37.47 & 0.07 & 0.00 & 0.07 & 35.67 & 0.06 \\ 
NaVid                   &   & Museum 5F R2 & 0.00 & 0.00 & 43.74 & 0.01 & 0.00 & 0.00 & 43.82 & 0.01 \\
Proposed                &   & Museum 5F R2 & \textbf{0.05} & 0.18 & 23.17 & 0.25 & 0.00 & \textbf{0.02} & 29.08 & \textbf{0.37} \\
Proposed                & ✓ & Museum 5F R2 & \textbf{0.05} & \textbf{0.32} & \textbf{16.43} & \textbf{0.29} & 0.00 & \textbf{0.02} & \textbf{24.78} & \textbf{0.37} \\
\midrule 
NavGPT                  &   & Museum 7F R1 & 0.00 & 0.00 & 54.70 & 0.08& 0.00 & 0.00 & 36.00 & 0.14 \\ %LAST
NaVid                   &   & Museum 7F R1 & 0.00 & 0.00 & 73.76 & 0.06 & 0.00 & 0.00 & 71.30 & 0.07 \\
Proposed                &   & Museum 7F R1 & \textbf{0.02} & \textbf{0.02} & 55.99 & 0.31 & 0.05 & 0.05 & 46.67 & 0.42 \\
Proposed                & ✓ & Museum 7F R1 & \textbf{0.02} & \textbf{0.02} & \textbf{42.59} & \textbf{0.48} & \textbf{0.09} & \textbf{0.09} & \textbf{25.22} & \textbf{0.67} \\
\midrule 
NavGPT                  &   & Museum 7F R2 & 0.00 & 0.00 & 61.64 & 0.01 &0.00 & 0.00 & 60.43 & 0.01 \\ %TO UPDATE
NaVid                   &   & Museum 7F R2 & 0.00 & 0.00 & 67.39 & 0.02 & 0.00 & 0.00 & 69.18 & 0.00 \\ %TO UPDATE
Proposed                &   & Museum 7F R2 & \textbf{0.15} & 0.26 & 47.41 & 0.17 & 0.00 & 0.10 & 52.81 & 0.16\\ %TO UPDATE
Proposed                & ✓ & Museum 7F R2 & 0.07 & \textbf{0.35} & \textbf{36.78} & \textbf{0.25} & \textbf{0.02} & \textbf{0.12} & \textbf{46.74} & \textbf{0.22} \\ %TO UPDATE
\bottomrule
\end{tabular}%
}
\vspace{-1em}
\end{table}

\vspace{-1mm}
\subsection{Performance of the Proposed Method}
As shown in Table~\ref{tab:result}, our model outperforms NavGPT and NaVid.
The baselines' suboptimal performances can be attributed to two factors: their deviation from the correct direction, and their premature decision that they had reached the goal. 
This is due to the fact that the baselines refer to the environment at each navigation step with an LLM. 
For example, if NavGPT makes a mistake even once during this process, it will be challenging for the model to recover the agent back to the correct path. 
Additionally, NaVid tends to make unnecessary frequent turns after initially following the route correctly for several iterations, likely due to the longer sequence of turns and longer instructions in our benchmark, which NaVid was not trained to handle.
In contrast, our method achieves the desired outcome through a single iteration of code generation inference, removing the need to initiate inferences at every intermediate step for instructions like \textit{``go straight for 100m and then turn right.''}.
We also observed that the instruction parsing module boosted the performance of our method in most cases.

\subsection{Difficulty of the Benchmark}
\label{sec:difficulty}
In Table~\ref{tab:result}, it is observed that the performances from onsite \revise{memory-based} instructions tended to be lower than those from online \revise{think-out-loud} instructions, as it is more likely for the route instructions to contain errors due to human memory, and it is harder for the system to recover from errors. 
Overall, our results demonstrate the difficulty of the instruction data from human memory and the value of our benchmark.

Across all routes, both our model and the baselines showed suboptimal or low performance. 
One major reason was the difficulty in handling the varied and inaccurate input instructions. 
In longer routes, participants tended to inaccurately estimate lengths for certain path segments and not include sufficient information about the destination.
\red{
Also, because our baseline contained modularized perception and control modules to focus on language parsing and reasoning capabilities, its suboptimal performance implies that the primary challenge in our benchmark lies in the complexity of the language instructions, such as inaccuracies or variations in wording, which were not present in previous benchmarks.
}
Upon closer inspection, many instructions in our benchmark contained phrases that required a combined understanding of both natural language and the building's structure, which our proposed model failed to follow.
One example was a phrase such as \textit{``go along this path and turn right in the first intersection,''} which was often described at the starting point of University R1. 
The instruction skips the right turn in the first turning point by describing it as \textit{``go along this path,''} because the building structure only allows a right turn at the immediate corner. 
As a result, the instruction starts by describing the first left turn where there are two possible directions to proceed.
This variation in the levels of topological details further highlights the difficulty of our benchmark, which imitates real-world scenarios of blind people seeking navigation instructions.

\red{
Furthermore, we realized that in the real-world, humans may be able to correct errors in the instructions. For example, some passersby instructed the robot to take a nonexistent corridor between the hexagon and the rectangular exhibitions near the museum 5F R2. Although the corridor did not exist, imagining its intended destination allowed locating an alternative route. Similarly, when participants provided incorrect turns, landmarks described later in the instructions helped identify and correct these oversights.
}
\begin{table}[]
\centering
\caption{\textbf{Effect of Instruction Refinement.} While in most cases refining instruction leads to an increase in performance, in certain cases, it was not always the case, due to redundant referral to surrounding objects.}
\label{tab:abalation}
\resizebox{\columnwidth}{!}{%
\begin{tabular}{cccccccccc}
\toprule
\multicolumn{2}{c}{Condition} & \multicolumn{4}{c}{\revise{Online Think-Out-Loud Instructions}} & \multicolumn{4}{c}{\revise{Onsite Memory-Based Instructions}} \\
Route & Iteration & SR↑  & OSR↑ & SPD↓  & CLS↑ & SR↑  & OSR↑ & SPD↓   & CLS↑ \\
\cmidrule(lr){1-2} \cmidrule(lr){3-6} \cmidrule(lr){7-10} 
University R1     & 1         & \textbf{0.43} & \textbf{0.48} & \textbf{16.37} & \textbf{0.56} & 0.25 & \textbf{0.40} & 20.41  & 0.52 \\
University R1     & 2         & 0.17 & 0.22 & 22.94 & 0.41 & \textbf{0.35} & 0.35 & \textbf{16.23}  & \textbf{0.56} \\
\midrule
University R2     & 1         & \textbf{0.04} & \textbf{0.04} & 86.82 & 0.36 & 0.00 & 0.00 & \textbf{93.32}  & \textbf{0.31} \\
University R2     & 2         & \textbf{0.04} & \textbf{0.04} & \textbf{76.36} & \textbf{0.41} & \textbf{0.05} & \textbf{0.05} & 102.95 & 0.28 \\
\midrule
Museum 5F R1     & 1         & \textbf{0.26} & \textbf{0.30} & \textbf{25.80} & 0.60 & 0.00 & 0.00 & \textbf{27.21} & \textbf{0.56}\\
Museum 5F R1     & 2         & 0.13 & 0.22 & 27.61 & \textbf{0.61} & \textbf{0.10} & \textbf{0.14} & 31.07 & 0.52\\
\midrule
Museum 5F R2     & 1         & \textbf{0.05} & 0.27 & 17.77 & 0.27 & 0.00 & 0.00 & 25.75 & 0.32 \\  %TO BE UPDATED
Museum 5F R2     & 2         & \textbf{0.05} & \textbf{0.36} & \textbf{15.09} & \textbf{0.30} & 0.00 & \textbf{0.05} & \textbf{23.82} & \textbf{0.43}\\  %TO BE UPDATED
\midrule
Museum 7F R1     & 1         & 0.00 & 0.00 & 46.57 & 0.43 & \textbf{0.09} & \textbf{0.09} & \textbf{23.77} & \textbf{0.68} \\
Museum 7F R1     & 2         & \textbf{0.05} & \textbf{0.05} & \textbf{38.61} & \textbf{0.53} & \textbf{0.09} & \textbf{0.09} & 26.66 & 0.65 \\
\midrule
Museum 7F R2     & 1         & 0.00 & 0.26 & 37.24 & \textbf{0.26} & 0.00 & 0.05 & \textbf{46.71} & 0.21 \\  %TO BE UPDATED
Museum 7F R2     & 2         & \textbf{0.13} & \textbf{0.43} & \textbf{36.33} & 0.24 & \textbf{0.05} & \textbf{0.19} & 46.76 & \textbf{0.23} \\  %TO BE UPDATED
\bottomrule
\end{tabular}%
}
\vspace{-1.5em}
\end{table}

% \begin{table}[]
% \centering
% \caption{Aggregated result}
% \label{tab:abalation}
% \resizebox{\columnwidth}{!}{%
% \begin{tabular}{ccccccccc}
% \toprule
% \multicolumn{1}{c}{} & \multicolumn{4}{c}{Online Study Data} & \multicolumn{4}{c}{Onsite Study Data} \\
% Iteration & SR↑  & OSR↑ & SPD↓  & CLS↑ & SR↑  & OSR↑ & SPD↓   & CLS↑ \\
% \cmidrule(lr){1-1} \cmidrule(lr){2-5} \cmidrule(lr){6-9} 
% 1 & \textbf{0.13 }& \textbf{0.23 }& 38.43 & 0.41 & 0.06 & 0.09 & \textbf{39.53} & 0.43 \\
% 2 & 0.10 & 0.22 & \textbf{36.16} & \textbf{0.42} & \textbf{0.11}& \textbf{0.15 }& 41.25 & \textbf{0.45} \\
% \bottomrule
% \end{tabular}%
% }
% \vspace{-1.5em}
% \end{table}
\subsection{Effect of Refining Instruction}
Table~\ref{tab:abalation}, reports the performance of our proposed method across different instruction iterations. 
The first iteration corresponds to the most natural, memory-based instruction, while the second represents memory-based instructions that are more accurate and contain features that may better assist the VLN agent in navigation. 
Generally, we found that refining the instruction led to a slight performance improvement. 
This suggests that, when deploying VLN-equipped robots, it is beneficial to assist sighted passersby in recalling routes more and in conveying environmental information in a format that is more compatible with robotic interpretation.
However, for Museum 5F R1 and University R1, the tendency was not always the case.
This happened because participants tended to describe more objects during the second iteration, which contained greater variation in object descriptions.
For example, one participant described only turning point-related information at the first iteration, while in the second iteration, the participant also described objects to ignore, such as, \textit{``then, you will come across an intersection with a door on the left and an intersection with doors on both sides, but ignore them and continue straight ahead.''}

\section{Conclusion and Future Work}
\label{sec:conclusion}
This work proposed Memory-Maze.
We found that realistic instructions collected in the onsite environment, where participants had to rely on human memories, were longer with greater variation in words, and contained more errors compared to the instructions collected online.
Upon qualitative inspection, we observed evidence of the tendency for \revise{memory-based} instruction to be more difficult for the model to handle, such as the ones that required understanding of \textit{``go along this path.''} 
This suggests that future VLN models should consider a more adaptive map representation where nodes and turns are not strictly defined, or a more flexible approach to accommodate varying topological descriptions.

\red{
For future work, we aim to explore the interactive aspect with users and robots.
For example, the robot could also guide the instruction from passersby to be better or rephrase it itself, potentially leading to better performance.
We also plan to convert our baseline into a closed-loop architecture that continuously verifies and refines its interpretation of instructions using real-time sensor input. 
}

\red{
Lastly, although the size of our benchmark is comparable to existing benchmarks~\cite{fu2025video,krantz2023iterative}, its size remains limited in order to be used as a dataset for training.
One possible approach is to modify the design of the online study so that annotators can only observe the environment prior to providing annotations, but this would only partially replicate the characteristics of real-world memory-based data.
Another potential method is to leverage LLMs with in-context learning to augment the benchmark.
However, further investigation is needed to ensure that LLM-generated data can accurately mimic the characteristics of our dataset.
}

\section*{ACKNOWLEDGEMENT}
\revise{
We deeply thank Hironobu Takagi for engaging in the discussion of this project.
We also thank participants in our experiment, from Carnegie Mellon University, Waseda University and Miraikan.
}

% \addtolength{\textheight}{-12cm}   

% \input{sections/98_Appendix}
% \input{sections/99_Acknowledgement}

\bibliographystyle{IEEEtran}
\bibliography{IEEEabrv}

@inproceedings{ranganeni2023exploring,
    author = {Ranganeni, Vinitha and Sinclair, Mike and Ofek, Eyal and Miller, Amos and Campbell, Jonathan and Kolobov, Andrey and Cutrell, Edward},
    booktitle = {HRI'23},
    title = {Exploring Levels of Control for a Navigation Assistant for Blind Travelers},
    year = {2023}
}

@inproceedings{krantz2023iterative,
  title={Iterative vision-and-language navigation},
  author={Krantz, Jacob and Banerjee, Shurjo and Zhu, Wang and Corso, Jason and Anderson, Peter and Lee, Stefan and Thomason, Jesse},
  booktitle={CVPR},
  year={2023}
}

@inproceedings{song2025towards,
  title={Towards long-horizon vision-language navigation: Platform, benchmark and method},
  author={Song, Xinshuai and Chen, Weixing and Liu, Yang and Chen, Weikai and Li, Guanbin and Lin, Liang},
  booktitle={CVPR},
  year={2025}
}

@inproceedings{fu2025video,
  title={Video-mme: The first-ever comprehensive evaluation benchmark of multi-modal llms in video analysis},
  author={Fu, Chaoyou and Dai, Yuhan and Luo, Yongdong and Li, Lei and Ren, Shuhuai and Zhang, Renrui and Wang, Zihan and Zhou, Chenyu and Shen, Yunhang and Zhang, Mengdan and others},
  booktitle={CVPR},
  year={2025}
}

@inproceedings{anderson2018vision,
    author = {Anderson, Peter and Wu, Qi and Teney, Damien and Bruce, Jake and Johnson, Mark and S{\"u}nderhauf, Niko and Reid, Ian and Gould, Stephen and Van Den Hengel, Anton},
    booktitle = {CVPR},
    title = {Vision-and-language Navigation: Interpreting Visually-grounded Navigation Instructions in Real Environments},
    year = {2018}
}

@inproceedings{chen2019touchdown,
    author = {Chen, Howard and Suhr, Alane and Misra, Dipendra and Snavely, Noah and Artzi, Yoav},
    booktitle = {CVPR},
    title = {Touchdown: Natural Language Navigation and Spatial Reasoning in Visual Street Environments},
    year = {2019}
}

@inproceedings{Dosovitskiy17,
    author = {Dosovitskiy, Alexey and Ros, German and Codevilla, Felipe and Lopez, Antonio and Koltun, Vladlen},
    booktitle = {CoRL},
    title = {CARLA: An Open Urban Driving Simulator},
    year = {2017}
}

@inproceedings{guerreiro2019cabot,
    author = {Guerreiro, Jo{\~a}o and Sato, Daisuke and Asakawa, Saki and Dong, Huixu and Kitani, Kris M and Asakawa, Chieko},
    booktitle = {ASSETS},
    title = {CaBot: Designing and Evaluating an Autonomous Navigation Robot for Blind People},
    year = {2019}
}

@inproceedings{huang2022language,
    author = {Huang, Wenlong and Abbeel, Pieter and Pathak, Deepak and Mordatch, Igor},
    booktitle = {ICML},
    title = {Language Models as Zero-shot Planners: Extracting Actionable Knowledge for Embodied Agents},
    year = {2022}
}

@inproceedings{huang2023visual,
    author = {Huang, Chenguang and Mees, Oier and Zeng, Andy and Burgard, Wolfram},
    booktitle = {ICRA},
    title = {Visual Language Maps for Robot Navigation},
    year = {2023}
}

@inproceedings{krantz2021waypoint,
    author = {Krantz, Jacob and Gokaslan, Aaron and Batra, Dhruv and Lee, Stefan and Maksymets, Oleksandr},
    booktitle = {CVPR},
    title = {Waypoint Models for Instruction-guided Navigation in Continuous Environments},
    year = {2021}
}

@inproceedings{kuribayashi2023pathfinder,
    author = {Kuribayashi, Masaki and Ishihara, Tatsuya and Sato, Daisuke and Vongkulbhisal, Jayakorn and Ram, Karnik and Kayukawa, Seita and Takagi, Hironobu and Morishima, Shigeo and Asakawa, Chieko},
    booktitle = {CHI},
    title = {PathFinder: Designing a Map-less Navigation System for Blind People in Unfamiliar Buildings},
    year = {2023}
}

@inproceedings{qi2020reverie,
    author = {Qi, Yuankai and Wu, Qi and Anderson, Peter and Wang, Xin and Wang, William Yang and Shen, Chunhua and Hengel, Anton van den},
    booktitle = {CVPR},
    title = {Reverie: Remote Embodied Visual Referring Expression in Real Indoor Environments},
    year = {2020}
}

@inproceedings{shah2023lm,
    author = {Shah, Dhruv and Osi{\'n}ski, B{\l}a{\.z}ej and Levine, Sergey and others},
    booktitle = {CoRL},
    title = {Lm-nav: Robotic Navigation with Large Pre-trained Models of Language, Vision, and Action},
    year = {2023}
}

@inproceedings{zhou2023navgpt,
    author = {Zhou, Gengze and Hong, Yicong and Wu, Qi},
    booktitle = {AAAI},
    title = {Navgpt: Explicit Reasoning in Vision-and-language Navigation with Large Language Models},
    year = {2024}
}

@inproceedings{biggie2023tell,
    author={Biggie, Harel and Mopidevi, Ajay Narasimha and Woods, Dusty and Heckman, Christoffer},
    title={Tell me where to go: A composable framework for context-aware embodied robot navigation},
    booktitle={CoRL},
    year={2023}
}

@inproceedings{gu2022vision,
  title={Vision-and-Language Navigation: A Survey of Tasks, Methods, and Future Directions},
  author={Gu, Jing and Stefani, Eliana and Wu, Qi and Thomason, Jesse and Wang, Xin},
  booktitle={ACL},
  year={2022}
}

@article{ilharco2019general,
    author = {Ilharco, Gabriel and Jain, Vihan and Ku, Alexander and Ie, Eugene and Baldridge, Jason},
    journal = {arXiv},
    title = {General Evaluation for Instruction Conditioned Navigation Using Dynamic Time Warping},
    year = {2019}
}

@inproceedings{jain2019stay,
    author = {Jain, Vihan and Magalhaes, Gabriel and Ku, Alexander and Vaswani, Ashish and Ie, Eugene and Baldridge, Jason},
    booktitle={ACL},
    title = {Stay on the Path: Instruction Fidelity in Vision-and-language Navigation},
    year = {2019}
}

@inproceedings{ku2020room,
  title={Room-Across-Room: Multilingual Vision-and-Language Navigation with Dense Spatiotemporal Grounding},
  author={Ku, Alexander and Anderson, Peter and Patel, Roma and Ie, Eugene and Baldridge, Jason},
  booktitle={EMNLP},
  year={2020}
}

@article{liu2023dragon,
  title={Dragon: A dialogue-based robot for assistive navigation with visual language grounding},
  author={Liu, Shuijing and Hasan, Aamir and Hong, Kaiwen and Wang, Runxuan and Chang, Peixin and Mizrachi, Zachary and Lin, Justin and McPherson, D Livingston and Rogers, Wendy A and Driggs-Campbell, Katherine},
  journal={RA-L},
  year = {2024}
}

@inproceedings{liu2023grounding,
    author = {Liu, Shilong and Zeng, Zhaoyang and Ren, Tianhe and Li, Feng and Zhang, Hao and Yang, Jie and Li, Chunyuan and Yang, Jianwei and Su, Hang and Zhu, Jun and others},
    booktitle={ECCV},
    title = {Grounding Dino: Marrying Dino with Grounded Pre-training for Open-set Object Detection},
    year = {2024}
}

@article{muller2022traveling,
    author = {M\"{u}ller, Karin and Engel, Christin and Loitsch, Claudia and Stiefelhagen, Rainer and Weber, Gerhard},
    journal = {TACCESS},
    title = {Traveling More Independently: A Study on the Diverse Needs and Challenges of People with Visual or Mobility Impairments in Unfamiliar Indoor Environments},
    year = {2022}
}

@inproceedings{suris2023vipergpt,
    author = {Sur{\'\i}s, D{\'\i}dac and Menon, Sachit and Vondrick, Carl},
    booktitle = {ICCV},
    title = {Vipergpt: Visual Inference Via Python Execution for Reasoning},
    year = {2023}
}

@article{wu2023vision,
    author = {Wu, Wansen and Chang, Tao and Li, Xinmeng and Yin, Quanjun and Hu, Yue},
    journal = {NCAA},
    title = {Vision-language Navigation: A Survey and Taxonomy},
    year = {2023},
}

@article{zhang2024navid,
    author = {Zhang, Jiazhao and Wang, Kunyu and Xu, Rongtao and Zhou, Gengze and Hong, Yicong and Fang, Xiaomeng and Wu, Qi and Zhang, Zhizheng and Wang, He},
    journal = {RSS},
    title = {NaVid: Video-based VLM Plans the Next Step for Vision-and-Language Navigation},
    year = {2024}
}

@inproceedings{kuribayashi2025wanderguide,
    title={WanderGuide: Indoor Map-less Robotic Guide for Exploration by Blind People},
    author={Kuribayashi, Masaki and Uehara, Kohei and Wang, Allan and Morishima, Shigeo and Asakawa, Chieko},
    booktitle = {CHI},
    year = {2025}
}

\end{document}